%% file: main.tex
\documentclass[sigconf]{acmart}
\makeatletter
\def\@ACM@checkaffil{
    \if@ACM@instpresent\else
    \ClassWarningNoLine{\@classname}{No institution present for an affiliation}%
    \fi
    \if@ACM@citypresent\else
    \ClassWarningNoLine{\@classname}{No city present for an affiliation}%
    \fi
    \if@ACM@countrypresent\else
        \ClassWarningNoLine{\@classname}{No country present for an affiliation}%
    \fi
}
\makeatother

\AtBeginDocument{%
  \providecommand\BibTeX{{%
    \normalfont B\kern-0.5em{\scshape i\kern-0.25em b}\kern-0.8em\TeX}}}

\setcopyright{acmlicensed}
\copyrightyear{2018}
\acmYear{2018}
\acmDOI{XXXXXXX.XXXXXXX}

\acmConference[Conference acronym 'XX]{Make sure to enter the correct
  conference title from your rights confirmation emai}{June 03--05,
  2018}{Woodstock, NY}
\acmISBN{978-1-4503-XXXX-X/18/06}

\usepackage{subfigure}
\usepackage{graphicx}
\usepackage{bbm}
\usepackage{multirow}
\usepackage{bbding}
\usepackage{amsmath}

\usepackage{algorithm}
\usepackage{algorithmic}
\usepackage{amsmath}

\usepackage{color}

\definecolor{light-gray}{gray}{0.80}

\usepackage{colortbl}

\usepackage{CJKutf8}

\begin{document}

\title{BuffGraph: Enhancing Class-Imbalanced Node Classification via Buffer Nodes}

\author{Qian Wang}
\affiliation{%
  \institution{National University of Singapore}
}
\email{qiansoc@nus.edu.sg}

\author{Zemin Liu}
\affiliation{%
  \institution{National University of Singapore}
}
\email{liu.zemin@hotmail.com}

\author{Zhen Zhang}
\affiliation{%
 \institution{National University of Singapore}
}
\email{zhen@nus.edu.sg}

\author{Bingsheng He}
\affiliation{%
  \institution{National University of Singapore}
}
\email{hebs@comp.nus.edu.sg}

\renewcommand{\shortauthors}{Trovato and Tobin, et al.}

\begin{abstract}
Class imbalance in graph-structured data, where minor classes are significantly underrepresented, poses a critical challenge for Graph Neural Networks (GNNs). To address this challenge, existing studies generally generate new minority nodes and edges connecting new nodes to the original graph to make classes balanced. However, they do not solve the problem that majority classes still propagate information to minority nodes by edges in the original graph which introduces bias towards majority classes. To address this, we introduce BuffGraph, which inserts buffer nodes into the graph, modulating the impact of majority classes to improve minor class representation. Our extensive experiments across diverse real-world datasets empirically demonstrate that BuffGraph outperforms existing baseline methods in class-imbalanced node classification in both natural settings and imbalanced settings. Code is available at https://anonymous.4open.science/r/BuffGraph-730A.
\end{abstract}

\ccsdesc[500]{Computing methodologies~Learning latent representations}
\ccsdesc[500]{Mathematics of computing~Graph algorithms}

\keywords{node classification; class imbalance; heterophily; graph neural network}


\received{20 February 2007}
\received[revised]{12 March 2009}
\received[accepted]{5 June 2009}


\maketitle

\input{sec-introduction}
\input{sec-preliminary}

\input{sec-related-work}

\input{sec-methods}
\input{sec-experiments}
\input{sec-conclusion}

\newpage
\bibliographystyle{ACM-Reference-Format}
\balance
\bibliography{sample-base}

\appendix
\input{sec-appendix}

\end{document}

%% file: sec-introduction.tex
\section{Introduction}

Graphs are powerful tools for representing complex data and relationships, as they can effectively encapsulate vast amounts of information \citep{cao2016deep}. Among various methods for graph analytics, graph representation learning stands out for its ability to convert high-dimensional graph data into a lower-dimensional space while retaining essential information \citep{cai2018comprehensive}. Recent years have witnesses the rapid development of graph representations, especially the Graph Neural Networks (GNN) \cite{kipf2016semi, velivckovic2017graph, hamilton2017inductive}. Among tasks on graphs, node classification has been important in real-world graphs such as classifying influencers in the social network \cite{bhadra2023graph} and detecting fraudsters in financial activities \cite{lin2022ethereum, shamsi2022chartalist}. 


Typically, GNN models are designed to operate under the assumption of class label balance, learning node features through neighborhood aggregation and presuming node similarity within these neighborhoods \cite{kipf2016semi, velivckovic2017graph, hamilton2017inductive}. However, this assumption does not hold across many real-world graphs, which often exhibit significant class imbalances. For example, the number of frauster accounts in the Ethereum network is much smaller than normal accounts \cite{wang2023etgraph}. Such class imbalances lead to a skewed performance between majority and minority classes, inherently biasing the the model towards majority classes \cite{zhou2023graphsr}. Consequently, directly applying GNNs to class-imbalanced graphs risks underrepresenting the minority classes, highlighting the need for approaches that can effectively address this imbalance.


To address this problem, multiple imbalanced node classification methods have been proposed \cite{zhao2021graphsmote, park2022graphens, song2022tam, zhou2023graphsr, li2023graphsha}. These methods can be categorized into two perspectives: data-level and algorithm-level. The data-level methods focus on balancing data distributions. For example, GraphSMOTE utilizes SMOTE to interpolate synthesize new minority nodes between two minority nodes in the embedding space \cite{zhao2021graphsmote}. GraphENS \cite{park2022graphens} resolves this problem by synthesizing new nodes by mixing minority and majority nodes. GraphSR augments minority classes from unlabelled nodes in the graph \cite{zhou2023graphsr}. GraphSHA synthesizes harder minority samples to enlarge the minor decision boundaries \cite{li2023graphsha}. In contrast, algorithm-level method TAM \cite{song2022tam} optimizes GraphENS by adjusting the margin of each minority node based on its neighborhood. 

\begin{figure*}[ht]
\centering
\includegraphics[width=0.49\linewidth]{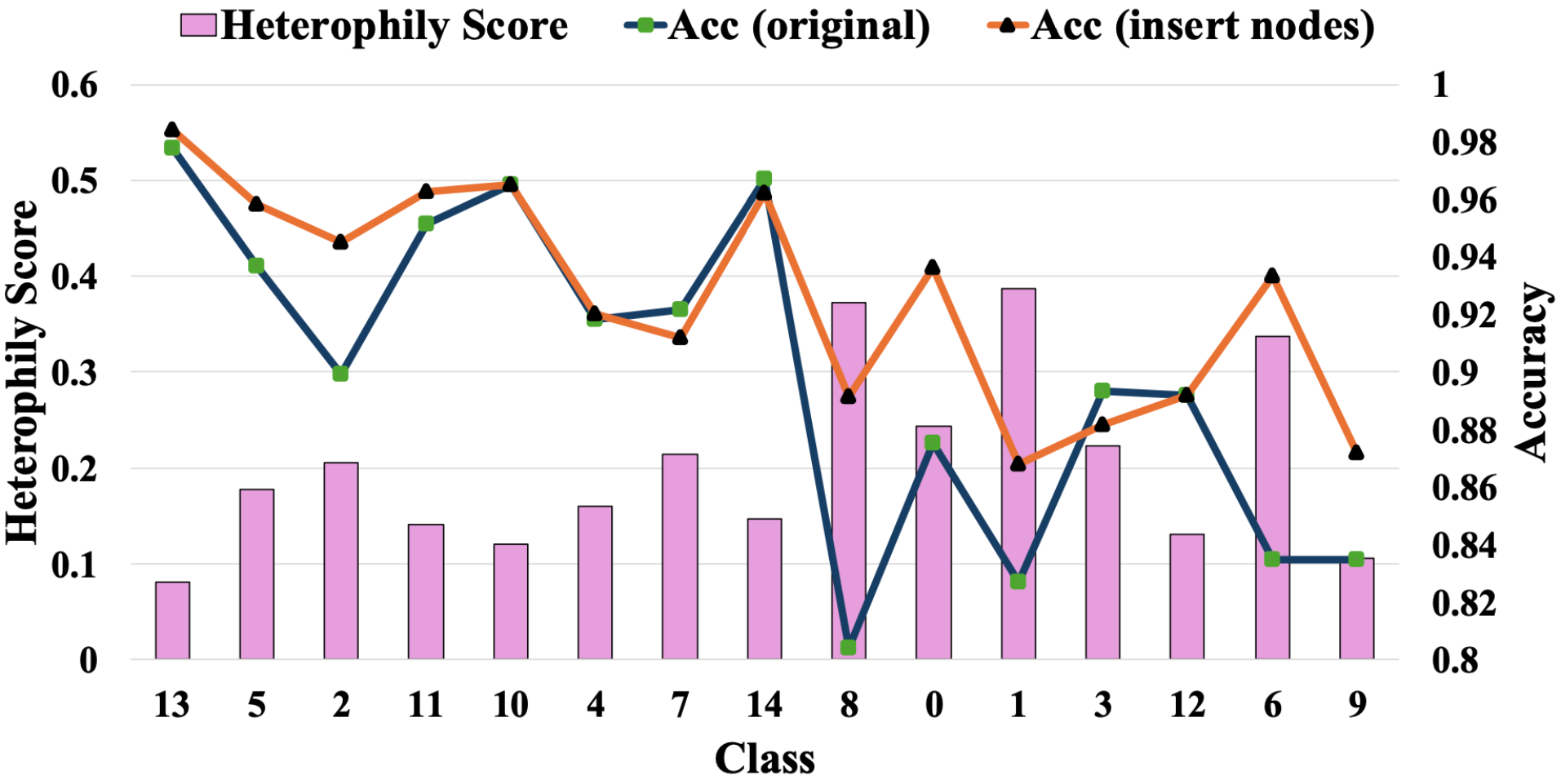}
\hfill 
\includegraphics[width=0.49\linewidth]{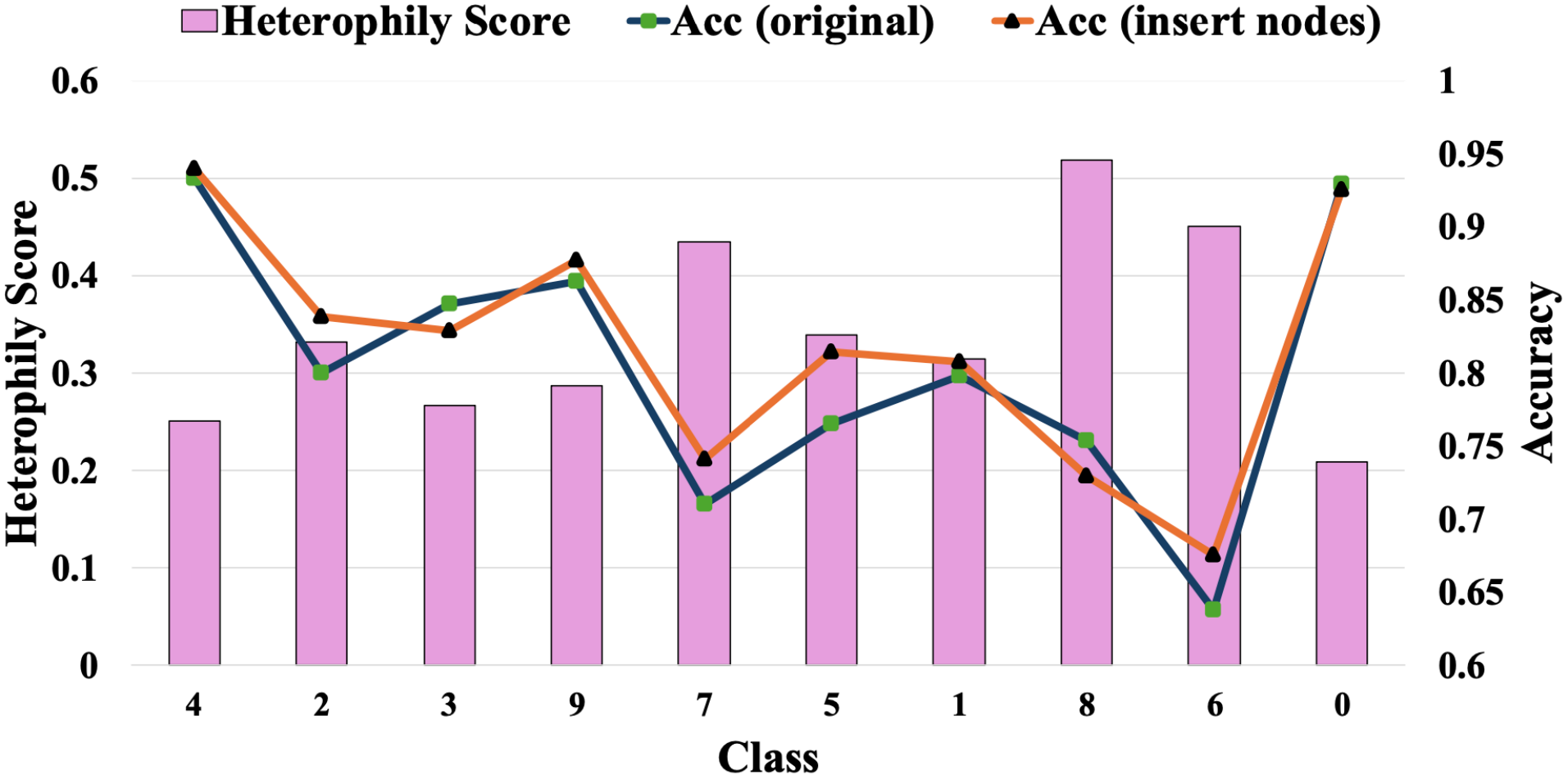}
\caption{Comparative analysis of class-wise accuracy improvements before and after node insertion for the Coauthor-CS (left) and WikiCS (right) datasets. Classes are organized in descending order according to the number of samples per class. The heterophily score depicted corresponds to the average heterophily score across samples within each class, offering insights into the impact of node insertion on class-wise performance considering the underlying heterophily dynamics.}
\label{fig:motivation}
\end{figure*}

Despite their effectiveness, these methods predominantly address imbalance from a node generation perspective, without directly tackling the critical issue of heterophily — the tendency for nodes to connect across different classes, which significantly influences performance. As depicted in Figure \ref{fig:motivation}, those minority classes with higher heterophily present a worse node classification accuracy. The primary limitation of prior approaches is their inability to effectively manage heterophily within their frameworks, thereby constraining their overall performance in real-world applications where heterophily is prevalent.

In this study, we address the challenge of class-imbalanced node classification by focusing on mitigating heterophily, a critical factor that significantly impacts the predictive performance for minority classes. Heterophily, the phenomenon where edges predominantly link nodes of different classes, presents a unique challenge in the realm of GNN particularly due to the failure of message passing mechanisms when adjacent nodes belong to distinct classes. This issue is compounded in scenarios characterized by a stark imbalance between major and minor classes, raising the question: how can we effectively address heterophily in such contexts to enhance classification outcomes? To answer this question, we hypothesize that instead of sampling nodes and edges subjectively, inserting buffer nodes into each edge can improve the performance of class-imbalanced node classification. And we show the performance change after inserting buffer nodes of two datasets: CoauthorCS and WikiCS in the Figure \ref{fig:motivation}. Heterophily is determined by analyzing the proportion of a node's one-hop neighbors that share the different labels as the node itself. Here we introduce the heterophily score of each class's definition: the average of heterophily of each node in this class. As illustrated in Figure \ref{fig:motivation}, the insertion of buffer nodes results in a noticeable improvement in accuracy across most classes especially those classes have a higher heterophily score.

Informed by our preceding experiments and the critical need to navigate the challenges posed by heterophily in graph structures, especially within the realm of class-imbalanced node classification, we have developed BuffGraph, an innovative model tailored to counteract heterophily's adverse effects. BuffGraph aims to improve prediction accuracy across both majority and minority classes comprehensively. Central to the BuffGraph methodology is the nuanced tailoring of message passing, which is delicately fine-tuned to match the heterophily level of each edge. BuffGraph distinguishes itself by employing buffer nodes to intelligently modulate the graph's heterophily. Buffer nodes are defined as artificial nodes inserted between existing nodes connected by an edge, serving as intermediaries that adjust the flow of information across the network. Buffer nodes are designed without labels, positioning them as neutral elements within the graph. This absence of labels allows buffer nodes to act purely as modulators of the information flow, without directly influencing the class-specific dynamics of message passing. Specifically, for edges characterized by high heterophily, BuffGraph predominantly routes message passing through these buffer nodes, effectively tempering the heterophily's impact. Conversely, for connections exhibiting low heterophily, it minimizes the intermediary role of buffer nodes, allowing for more direct message passing. Such adaptability ensures that BuffGraph adeptly addresses the distinct heterophily challenges presented by different classes, facilitating more precise and fair predictions within class-imbalanced graphs.


To rigorously evaluate the efficacy of our BuffGraph, we conduct experiments on a variety of real-world datasets, including co-purchase networks (AmazonPhotos and AmazonComputers \cite{shchur2018pitfalls}), co-authorship networks (CoauthorCS and CoauthorPhysics \cite{sen2008collective}), and a Wikipedia-based dataset (WikiCS \cite{mernyei2020wiki}). These datasets are well-acknowledged for their relevance and complexity, offering a robust framework for testing. Our experiment settings are designed to be comprehensive and rigorous, comparing BuffGraph's performance against several baseline models across natural splitting and imbalanced splitting settings. The results demonstrate that BuffGraph consistently outperforms the baseline models \cite{zhao2021graphsmote, park2022graphens, song2022tam, li2023graphsha} in most configurations and datasets. For instance, under the naturally splitting scenario, BuffGraph exhibits its superiority by achieving a 3\% increase in accuracy, a 2\% enhancement in balanced accuracy, and a significant 4\% boost in F1-score on the Amazon-Computers dataset compared to the second-best outcomes. Conversely, within the imbalanced splitting framework, BuffGraph continues to excel, marking a 2\% improvement in accuracy, a 1\% gain in balanced accuracy, and a 1.5\% uplift in F1-score on the Amazon-Computers dataset relative to the runner-up results. These findings underscore BuffGraph's adaptability and efficacy in addressing class imbalances across diverse real-world graph scenarios.

In summary, our work introduces several key contributions:

\begin{itemize}
    \item We propose the BuffGraph model, a novel approach ingeniously crafted to refine message passing for each edge in a graph. This model is specifically engineered to address the significant challenge of heterophily in scenarios of class imbalance during node classification, ensuring precise and effective communication between nodes.
    \item Through extensive experimental evaluations, BuffGraph is shown to outperform an array of baseline models in terms of performance and adaptability. This clearly establishes its superiority and benchmark setting capability in handling diverse graph data complexities.
    \item The insights derived from our study not only highlight the practical efficacy of the BuffGraph model but also pave new avenues for future explorations into the nuanced interplay of heterophily in class-imbalanced graph data analysis.
\end{itemize}

%% file: sec-preliminary.tex
\section{Preliminary}
In this section, we introduce the notations and definitions appearing in our BuffGraph model. 

\subsection{Graph Notations}

In this study, we address the challenge of class-imbalanced node classification on an unweighted, undirected graph $\mathcal{G} = (\mathcal{V}, \mathcal{E})$, where $\mathcal{V} = \{v_1, \cdots, v_N\}$ represents the set of $N$ nodes, and $\mathcal{E} \subseteq \mathcal{V} \times \mathcal{V}$ denotes the edges. The graph structure is captured by an adjacency matrix $\boldsymbol{A} \in \{0, 1\}^{N \times N}$, with $\boldsymbol{A}_{ij} = 1$ indicating an edge between nodes $v_i$ and $v_j$. Node features are represented by a matrix $\boldsymbol{X} \in \mathbb{R}^{N \times d}$, where each row $\boldsymbol{X}_i \in \mathbb{R}^d$ corresponds to the $d$-dimensional features of node $v_i$. Each node $v$ is labeled with one of $C$ classes, $\boldsymbol{Y}(v) \in \{1, \cdots, C\}$, with $\boldsymbol{Y}_c$ encompassing all nodes in class $c$.

The training subset $\mathcal{V}^L \subset \mathcal{V}$, particularly in imbalanced settings, is characterized by class size disparities, quantified by the imbalance ratio 
$\rho = \max \{|\boldsymbol{Y}_i|\}^{C}_{i=1} / \min \{|\boldsymbol{Y}_j|\}^{C}_{j=1}$, 
reflecting the size proportion between the largest and smallest classes. This setup underscores the complexities of achieving fair and accurate classification across diverse class distributions in graph-based learning scenarios.

\subsection{Node Classification via GNN}

GNN serve as a cornerstone for node classification within graph-structured data. The functionality of the \(l\)-th layer in GNNs is calculated through a set of operations: the message function \(m_l\), the aggregation function \(\psi_l\), and the update function \(\gamma_l\) \cite{park2022graphens}. The update process for a node \(v\)'s feature \(x_v^{(l+1)}\) from its \(l\)-th layer feature \(x_v^{(l)}\) is given by:
\begin{equation}
x_v^{(l+1)} = \gamma_l \left( x_v^{(l)}, \psi_l \left( \left\{ m_l\left(x_u^{(l)}, x_v^{(l)}, w_{u,v}\right) \mid u \in \mathcal{N}(v) \right\}\right)\right),
\end{equation}
where \(w_{u,v}\) represents the weight of the edge between nodes \(u\) and \(v\). For instance, GCN \cite{kipf2016semi} computes the node feature update as:
\begin{equation}
x_v^{(l+1)} = \sum_{u \in \mathcal{N}(v) \cup \{v\}} \frac{1}{\sqrt{\hat{d}_v \hat{d}_u}} \Theta_l x_u^{(l)},
\end{equation}
where \(\hat{d}_v\) denotes the normalized degree of node \(v\), incorporating the self-loop, and \(\Theta_l\) is the layer-specific trainable parameter matrix. 

\subsection{Heterophily Score}








In this work, we address the challenge of class-imbalanced node classification through the lens of heterophily. We propose the heterophily score as a pivotal metric in our analysis, designed to quantify the diversity of connections within the network. Specifically, it measures the degree to which nodes or edges deviate from the principle of homophily -- the tendency of similar nodes to be connected. This metric is computed for each node, edge, and class within the graph, facilitating a comprehensive understanding of the network's structure and its implications for node classification.

For nodes, we adopt the definition in \cite{pei2020geom} to define the heterophily score of a node \(v\) as:
\begin{equation}
\beta_v = 1 - \frac{|\{\boldsymbol{Y}(u)=\boldsymbol{Y}(v)\}_{u\in\mathcal{N}(v)}|}{|\mathcal{N}(v)|},
\end{equation}
where $\mathcal{N}(v)$ is the set of one-hop neighbors of node \(v\).
This equation calculates the proportion of a node's neighbors that do not share its label, thus quantifying the node's heterophily.

To address the limitations inherent in training data, where labels may not fully capture the diversity of connections, we also compute a heterophily score for each edge. This score aids in refining our predictive models by emphasizing the dynamics of relationships within the graph. The edge heterophily score, based on the Manhattan distance between the embeddings of two connected nodes \(u\) and \(v\), reflects the intuition that more similar embeddings indicate lower heterophily. The formula is as follows:
\begin{equation}
h_{uv} = \frac{1}{D} \sum_{i=1}^{D} |z_u^i - z_v^i|,
\end{equation}
where \(z_u\) and \(z_v\) are the embeddings of nodes \(u\) and \(v\) respectively, and \(D\) is the dimensionality of the embeddings.

Lastly, we calculate the class-level heterophily score as the average heterophily score of all nodes belonging to a class \(c\), offering a measure of the class's overall connectivity diversity:
\begin{equation}
h_{c} = \frac{1}{|C|} \sum_{v \in C} \beta_v,
\end{equation}
where \(C\) represents the set of nodes in class \(c\), and \(\beta_v\) is the heterophily score of node \(v\). Through this multi-faceted approach to computing heterophily scores at different structural levels of the graph, we gain a nuanced insight into how heterophily impacts imbalanced node classification.


%% file: sec-related-work.tex
\section{Related Work}
In this section, we review the prior work related to class-imbalanced node classification, and heterophily in graphs.



\subsection{Class-imbalanced Node Classification}
Class imbalance is a common phenomenon in machine learning tasks such as image classification \cite{peng2017object} and fake account detection \cite{he2009learning}. For class-imbalanced graphs, majority classes have more samples than minority classes, leading to a bias towards the majority. Existing methods to tackle class imbalance in graphs primarily involve generating new nodes and edges \cite{shi2020multi, qu2021imgagn, zhou2023graphsr, zhao2021graphsmote, park2022graphens, li2023graphsha}. Among these methods, DRGCN \cite{shi2020multi} employs GANs for synthetic node generation. GraphSMOTE \cite{zhao2021graphsmote} and ImGAGN \cite{qu2021imgagn} generate nodes belonging to minority classes to balance the classes. However, ImGAGN \cite{qu2021imgagn} focuses on binary classification, making it less suitable for multi-class graphs. GraphENS \cite{park2022graphens} generates minority nodes by considering the node's neighborhood and employing saliency-based mixing. TAM \cite{song2022tam} addresses class imbalance by introducing connectivity- and distribution-aware margins to guide the model, emphasizing class-wise connectivity and the distribution of neighbor labels. GraphSR \cite{zhou2023graphsr} augments minority classes from unlabelled nodes of the graph automatically. GraphSHA \cite{li2023graphsha} focuses on generating 'hard' minority classes to enlarge the margin between majority and minority classes. However, these methods do not address the heterophily phenomenon in class-imbalanced graphs, where majority classes disproportionately influence minority classes, further inducing bias towards the majority. 

\subsection{Heterophily in Graphs}
Although numerous GNNs have been proposed, most operate under the assumption that nodes with similar features or those belonging to the same class are connected to each other \cite{zheng2022graph}. For instance, a paper is likely to cite other papers from the same field \cite{sen2008collective}. However, this assumption does not hold for many real-world graphs. For instance, fraudulent accounts may connect to numerous legitimate accounts in financial datasets. Heterophily refers to the connection between two nodes with dissimilar features or from different classes \cite{zheng2022graph}. GNNs designed for heterophilic graphs primarily fall into two categories: non-neighbor message aggregation and GNN architecture refinement. MixHop \cite{abu2019mixhop} considers two-hop neighbors instead of one-hop. H2GCN \cite{zhu2020beyond} theoretically demonstrates that two-hop neighbors contain more nodes from the same class as the central node than one-hop neighbors do. In contrast, GNN architecture refinement focuses on distinguishing similar neighbors from dissimilar ones \cite{yan2022two, zhu2021graph}. Inspired by these methods, we propose adjusting the influence of minority nodes' neighbors based on the degree of heterophily between them and their one-hop neighbors in the context of class-imbalanced node classification tasks.

%% file: sec-methods.tex
\section{Proposed Method}

\begin{figure*}[!t]
\centering
\includegraphics[width=1.0\linewidth]{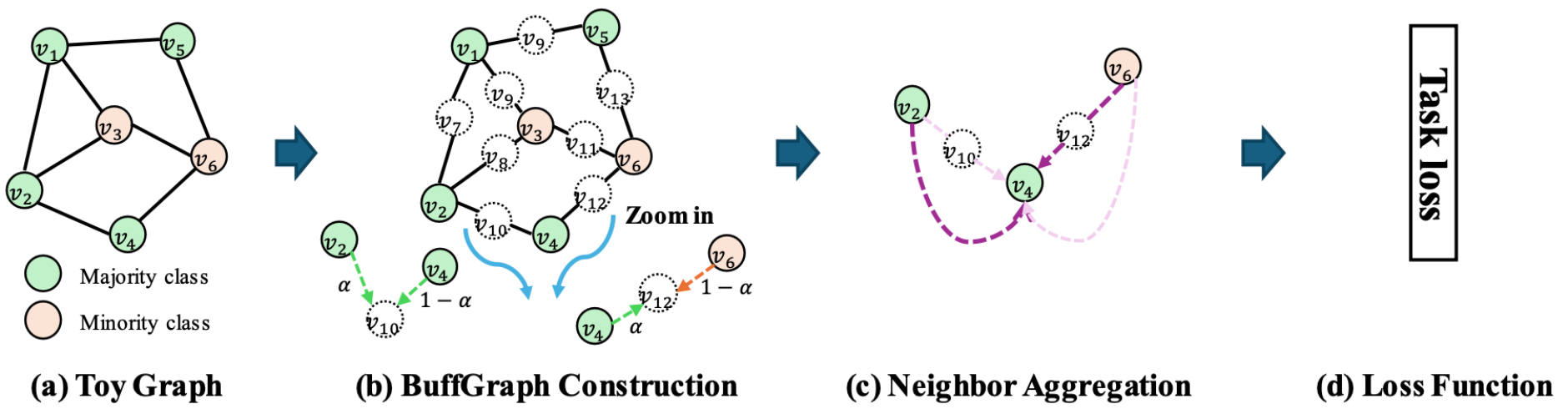}
\centering
\caption{
BuffGraph overview where $v_1$, $v_2$, $v_4$, $v_5$ are of the major class and $v_3$, $v_6$ are of the minor class. The input graph is shown in (a). Subsequently, we introduce a buffer node into every edge within the graph, as depicted in (b). The feature of the buffer node is a blend of the features from the two nodes connected by the edge, weighted by $\alpha$ and $1 - \alpha$ respectively.
We zoom in $v_4$ to show the neighbor aggregation of BuffGraph in (c). Each neighbor node passes message both across the buffer node and directly to $v_4$ at the same time based on the edge's heterophily extent. The loss function is calculated when doing the neighbor aggregation in BuffGraph shown in (d). }
\label{figure:framework}
\end{figure*}

In this section, we present a comprehensive description of our BuffGraph model. The framework of BuffGraph is depicted in the Figure \ref{figure:framework}. First, we insert buffer nodes into every edge in the graph. Then, we refine message passing and optimize neighbor aggregation under this setup, all guided by edge heterophily. And Algorithm \ref{algo:dynamicbuffernode} outlines the steps of BuffGraph's implementation. Additionally, we delve into both the complexity and theoretical analysis of BuffGraph, thereby substantiating its effectiveness from a theoretical standpoint. 

\subsection{Buffer Node Generation}

In response to the challenge posed by class imbalance in node classification tasks, we propose a pioneering buffer node generation approach, inspired by the mixup concept \cite{azabou2023half}. Buffer nodes introduce new nodes along each edge, thereby extending the path messages traverse, which effectively slows the message passing. This extension transforms direct one-hop neighbors into two-hop neighbors, broadening the interaction range and subtly modulating communication speed within the network. 

A buffer node does not have a label but have a feature vector. Specifically, the feature vector $\boldsymbol{X}_{\text{buf}}$ for a buffer node $v_{\text{buf}}$ is generated by interpolating the features of the two connected nodes: a source node $v_{\text{src}}$ and a target node $v_{\text{tar}}$. The interpolation is governed by the following equation:
\begin{equation}
\boldsymbol{X}_{\text{buf}} = \alpha \boldsymbol{X}_{\text{src}} + (1 - \alpha) \boldsymbol{X}_{\text{tar}}, \quad \alpha \in [0, 1].
\end{equation}
Here, $\alpha$ serves as the mixup coefficient, influencing the degree to which the features of the source and target nodes impact the buffer node's features. A lower value of $\alpha$ biases the feature vector $\boldsymbol{X}_{\text{buf}}$ towards the target node $v_{\text{tar}}$ while a higher value of $\alpha$ biases the feature vector $\boldsymbol{X}_{\text{buf}}$ towards the source node $v_{\text{src}}$.

The integration of a buffer node into each edge is designed primarily to modulate the message passing across edges, especially in contexts marked by heterophily, rather than expanding the feature set. Consequently, the role of $\alpha$ becomes secondary in this context, and we opt for a uniform value of $\alpha = 1/2$ for all edges to simplify the initial phase of our research and ensure methodological consistency. To rigorously assess the impact of $\alpha$, we undertake a series of experiments exploring variations of $\alpha$, which are detailed in Section~\ref{sec:parameter}. And to validate the efficacy of decelerating message passing speed through the insertion of buffer nodes, we present a theoretical analysis in Section~\ref{sec:theory}.

\subsection{BuffGraph Framework}


To address the challenges of message passing across heterophilic edges within class-imbalanced graphs, we introduce the BuffGraph model. This model incorporates a dynamic message passing mechanism that precisely modulates the flow of information through each edge, optimizing the efficacy of buffer nodes in managing heterophily. A detailed illustration of the BuffGraph framework and its operational intricacies is provided in Figure~\ref{figure:framework}.

Initially, we introduce an innovative approach to integrate buffer nodes into every edge of the graph, a strategy that diverges from the rigid node insertion method utilized in \cite{azabou2023half} to address heterophily in class-imbalanced graphs. Our method simultaneously employs both the original edge, acting as a residual link, and the newly established links through buffer node insertion. This dual-link utilization allows for dynamic modulation of the message passing process. Following this, we pre-train a conventional GCN model to derive a heterophily score for each edge. This score then guides the allocation of information flow, enabling a differential treatment of messages: they are passed directly via residual links or the links generated by buffer nodes, with weights assigned based on the heterophily score. For instance, as depicted in Figure~\ref{figure:framework}, node $v_4$ shares similarity with $v_2$ but differs from $v_6$. As a result, the message from $v_6$ predominantly traverses via buffer node $v_{12}$, whereas the message from $v_2$ has a lesser reliance on buffer node $v_{10}$. Two special cases are: one where only residual links are used for direct message passing, which is the vanilla GNN; another where only links generated by buffer nodes are employed for indirect message passing. The GNN neighbor aggregation mechanism we utilize is \cite{li2023graphsha}:
\begin{equation}
\boldsymbol{H}_{t}^{(l)} \leftarrow \boldsymbol{\operatorname { Transform }}\left(\underset{\forall v_s \in \mathcal{N}_t}{\boldsymbol{\operatorname { Propagate }}}\left(\boldsymbol{H}_{s}^{(l-1)} ; \boldsymbol{H}_{t}^{(l-1)}\right)\right),
\end{equation}
where $\boldsymbol{H}_{t}^{(l)}$ represents node embedding of node $v_t$ in the $l$-th layer.

Then, during the training phase, we leverage the message passing speed fine-tuned in the pre-training stage to further train the model. Notably, we reassess the heterophily score for each edge based on current model outputs and adjust the message passing speed accordingly by taking into account both heterophily loss and node classification loss every 50 epochs. The loss function is:
\begin{equation}
\boldsymbol{\mathcal{L}_{\text{total}}} = \boldsymbol{\mathcal{L}_{\text{pred}}} + \lambda \cdot \boldsymbol{\mathcal{L}_{\text{hetero}}}
\end{equation}

This adjustment is grounded in the understanding that edges linking nodes of differing classes (indicative of high heterophily) should minimize direct message passing in favor of routing messages through buffer nodes, and vice versa. 
By judiciously modulating the information flow in this manner, we enhances our model's flexibility and effectiveness across various graph structures and class imbalance situations. The detailed algorithm of our comprehensive framework is outlined in Algorithm~\ref{algo:dynamicbuffernode}.

\begin{algorithm}[t]
\caption{Buffer Node Synthesis and Adaptive Message Passing Algorithm}
\label{algo:dynamicbuffernode}
\flushleft{\textbf{Input}: Graph $\mathcal{G}=(\boldsymbol{X},\boldsymbol{A})$, training set nodes $\mathcal{V}^L$ and their labels $\boldsymbol{Y}^L$, number of classes $C$}

\textbf{Parameters}: Mixup coefficient $\alpha$, Heterophily loss coefficient $\lambda$ \\

\begin{algorithmic}[1] 
\STATE \textbf{Buffer Node Generation:}
\FOR{each edge $(v_{\text{src}}, v_{\text{tar}})$ in $\mathcal{G}$}
    \STATE Generate buffer node feature $\boldsymbol{X}_{\text{buf}}$ using Eq.~(1) with $\alpha$
    \STATE Insert buffer node $v_{\text{buf}}$ with feature $\boldsymbol{X}_{\text{buf}}$ into $\mathcal{G}$ between $v_{\text{src}}$ and $v_{\text{tar}}$
\ENDFOR

\STATE \textbf{Pre-training Stage:}
\STATE Initialize vanilla GNN $f_{\theta}$
\FOR{each node $v \in \mathcal{V}^L$}
    \STATE Pre-train $f_{\theta}$ for node classification
    \STATE Obtain node embeddings for all nodes in $\mathcal{G}$ to decide the message passing speed of each edge
\ENDFOR
\STATE Decide message passing for all edges in $\mathcal{G}$

\STATE \textbf{Training Stage:}
\FOR{epoch = 1 to $N$}
    \IF{epoch \% 50 == 0}
        \STATE Recompute heterophily scores for all edges in $\mathcal{G}$
    \ENDIF
    \FOR{each edge $(v_{i}, v_{j})$ in $\mathcal{G}$ including edges to buffer nodes}
        \STATE Adjust message passing speed based on heterophily scores
    \ENDFOR
    \STATE Compute node prediction loss $\mathcal{L}_{\text{pred}}$
    \STATE Compute heterophily loss $\mathcal{L}_{\text{hetero}}$
    \STATE Total loss $\mathcal{L} = \mathcal{L}_{\text{pred}} + \lambda \cdot \mathcal{L}_{\text{hetero}}$
    \STATE Update $f_{\theta}$ by minimizing $\mathcal{L}$ with standard backpropagation
\ENDFOR

\STATE \textbf{return} trained $f_{\theta}$
\end{algorithmic}
\end{algorithm}

\subsection{Complexity Analysis}
Given \( N \) as the total number of nodes and \( E \) as the total number of edges within the graph, the creation of buffer nodes correlates with the entirety of the node and edge sets, leading to a computational complexity of \( O(N + E) \). This process efficiently reuses mixup coefficients from previous epochs, thus avoiding additional computational demands. Pairing nodes for the mixup procedure exhibits a complexity of \( O(N^2) \), given the potential for any node to be paired with another. Generating augmented features for buffer nodes requires \( O(N \cdot d) \) time, with \( d \) representing the dimensionality of the node features. Additionally, forming augmented edges for the buffer nodes incurs a time complexity of \( O(E) \), engaging the whole graph's edge set. Hence, the total additional complexity introduced by our model is \( O(N^2 + N \cdot d + E) \). As demonstrated by successful experiments on extensive datasets such as Coauthor-Physics and WikiCS, BuffGraph can work well on the large datasets.

\subsection{Theoretical Analysis} \label{sec:theory}
In this section, we explore the BuffGraph model, with a special emphasis on understanding how the integration of buffer nodes alters the graph's structure through changes in eigenvalues \cite{azabou2023half}. We examine the pivotal role these modifications play in addressing the challenges associated with heterophily.

\noindent \textbf{Graph Laplacian and Eigenvalues in BuffGraph.}
For a graph $\mathcal{G} = (\mathcal{V}, \mathcal{E})$ with nodes $\mathcal{V}$ and edges $\mathcal{E}$, the graph Laplacian $\mathbf{L}$ is defined as $\mathbf{L} = \mathbf{D} - \mathbf{A}$, where $\mathbf{A}$ is the adjacency matrix and $\mathbf{D}$ is the degree matrix. The eigenvalues of $\mathbf{L}$, represented by $\lambda_i$ for $i = 1, \ldots, |\mathcal{V}|$, play a crucial role in the graph's diffusion processes. Specifically, a smaller eigenvalue for a node suggests that it is more easily influenced by its neighbors' diffusion patterns.

\noindent \textbf{Impact of Buffer Nodes on Spectral Properties.}
Buffer nodes alter the adjacency matrix to $\mathbf{A}'$ and degree matrix to $\mathbf{D}'$. This results in a new graph Laplacian $\mathbf{L}' = \mathbf{D}' - \mathbf{A}'$. The eigenvalues $\lambda'_i$ of $\mathbf{L}'$ now encapsulate the modified spectral characteristics, affecting the graph's information propagation behavior. Then, the eigenvalue shift as follows:
\begin{equation}
\Delta\lambda_i = \lambda'_i - \lambda_i,
\end{equation}
where $\Delta\lambda_i$ quantifies the change in spectral properties attributable to the addition of buffer nodes. With the increase of eigenvalues, a minority node preserves more of its magnitude without being affected by its majority neighbors \cite{keriven2022not}.

\noindent \textbf{Modified Message Passing in BuffGraph.}
The incorporation of buffer nodes necessitates a revision of the message passing mechanism, which in the spectral domain, is reflected by:
\begin{equation}
\mathbf{H}^{(l+1)} = \sigma\left(\mathbf{D}'^{-1/2} \mathbf{A}' \mathbf{D}'^{-1/2} \mathbf{H}^{(l)} \mathbf{W}^{(l)}\right),
\end{equation}
where $\mathbf{H}^{(l)}$ denotes node embeddings at layer $l$, and $\mathbf{W}^{(l)}$ is the weight matrix at layer $l$. This formula underscores the nuanced adaptation of the BuffGraph model to heterophily by modulating the eigenvalues through buffer nodes.

\noindent \textbf{Conclusion.}
Buffer nodes adjust the eigenvalues $\lambda'_i$ of the original graph to facilitate a more balanced diffusion of information, ensuring both minority and majority class features are adequately represented.

%% file: sec-experiments.tex
\section{Experiments}
In this section, we conduct extensive experiments to evaluate the effectiveness of BuffGraph for class-imbalanced node classification compared with existing baselines. We would like to answer the following three research questions:

\noindent \textbf{\textit{Q1:} How does BuffGraph's performance in node classification on naturally class-imbalanced graphs compare to that of existing baseline models?}

\noindent \textbf{\textit{Q2:} How effectively does BuffGraph outperform other baseline models in node classification across graphs with varying class-imbalance ratios?}

\noindent \textbf{\textit{Q3:} Do all of BuffGraph's components contribute to its performance?}

\begin{table}[!t]
\centering
\begin{center}
\caption{Statistics of datasets used in the paper.}\label{table:datasets}
\vspace{-0.2cm}
\scalebox{0.8}{
\begin{tabular}{l|ccccc}
\toprule
{Dataset}  & {\#Nodes} & {\#Edges} & {\#Features} & {\#Classes}& {\#Max / Min}\\
\midrule
Amazon-Photos  & 7,650 & 119,081 & 745 & 8 & 5.86 \\
Amazon-Computers  & 13,752 & 245,861 & 767 & 10 & 17.72 \\
Coauthor-CS  & 18,333 & 163,778 & 6,805 & 15 & 35.05\\
Coauthor-Physics & 34,493 & 247,962 & 8,415 & 5 & 6.33 \\
WikiCS & 11,701 & 216,123 & 300 & 10 & 9.08 \\
\bottomrule
\end{tabular}
}
\end{center}
\vspace{-0.2cm}
\end{table}

\begin{table*}[!t]
\centering
\begin{center}
\caption{Random splitting experiment results of BuffGraph and other baselines on five class-imbalanced node classification benchmark datasets. We report all metrics with the standard deviation errors for five repetitions. The best result is highlighted by bold text. The runner-up result is highlighted by the underline. }\label{table:main-results}
\vspace{-0.1cm}
\scalebox{0.65}{
\begin{tabular}{cl|ccc|ccc|ccc|ccc|ccc}
\toprule
\multirow{2}{*}{} & \textbf{Dataset} & \multicolumn{3}{c|}{Amazon-Photos} & \multicolumn{3}{c|}{Amazon-Computers} & \multicolumn{3}{c|}{Coauthor-CS} & \multicolumn{3}{c|}{Coauthor-Physics} & \multicolumn{3}{c}{WikiCS} \\
\cmidrule(lr){3-5}\cmidrule(lr){6-8}\cmidrule(lr){9-11}\cmidrule(lr){11-13}\cmidrule(lr){13-15}\cmidrule(lr){15-17}
&Random Splitting  & {Acc.} & {BAcc.} & {F1} & {Acc.} & {BAcc.} & {F1} & {Acc.} & {BAcc.} & {F1} & {Acc.} & {BAcc.} & {F1} & {Acc.} & {BAcc.} & {F1}  \\
\midrule
\multirow{11}{*}{\rotatebox{90}{\textbf{Methods}}}
& Vanilla & 92.44\footnotesize{$\pm$0.16} & 90.41\footnotesize{$\pm$0.62} & 91.20\footnotesize{$\pm$0.27} & 
\underline{87.71}\footnotesize{$\pm$0.36} & 82.34\footnotesize{$\pm$1.36} & 83.03\footnotesize{$\pm$1.86} & 
92.89\footnotesize{$\pm$0.41} & 89.97\footnotesize{$\pm$0.45} & 90.70\footnotesize{$\pm$0.63} &
96.22\footnotesize{$\pm$0.24} & 94.60\footnotesize{$\pm$0.42} & 93.49\footnotesize{$\pm$0.03} & 
83.20\footnotesize{$\pm$0.23} & 80.34\footnotesize{$\pm$0.41} & 80.63\footnotesize{$\pm$0.07}\\
\cmidrule(lr){2-17}
& Reweight & 92.91\footnotesize{$\pm$0.36} & 92.51\footnotesize{$\pm$0.41} & 91.72\footnotesize{$\pm$0.25} & 86.21\footnotesize{$\pm$0.71} & \underline{89.23}\footnotesize{$\pm$0.19} & 85.09\footnotesize{$\pm$0.68} & 
92.86\footnotesize{$\pm$0.03} & 90.86\footnotesize{$\pm$0.13} & 91.09\footnotesize{$\pm$0.04} &
95.70\footnotesize{$\pm$0.02} & 95.06\footnotesize{$\pm$0.05} & 94.52\footnotesize{$\pm$0.01} &
82.66\footnotesize{$\pm$0.17} & \underline{82.82}\footnotesize{$\pm$0.09} & 80.92\footnotesize{$\pm$0.20}\\
& PC Softmax & 91.83\footnotesize{$\pm$0.33} & 91.83\footnotesize{$\pm$0.34} & 90.31\footnotesize{$\pm$0.48} & 87.13\footnotesize{$\pm$1.49} & 87.60\footnotesize{$\pm$0.67} & 85.09\footnotesize{$\pm$1.71} & 
92.95\footnotesize{$\pm$0.12} & \underline{91.87}\footnotesize{$\pm$0.11} & 91.21\footnotesize{$\pm$0.11} &
96.14\footnotesize{$\pm$0.07} & 95.36\footnotesize{$\pm$0.11} & 95.07\footnotesize{$\pm$0.11} &
82.76\footnotesize{$\pm$0.32} & 81.94\footnotesize{$\pm$0.50} & 80.46\footnotesize{$\pm$0.45} \\
& Balanced Softmax & 93.46\footnotesize{$\pm$0.05} & 91.19\footnotesize{$\pm$0.01} & 92.05\footnotesize{$\pm$0.05} & 86.28\footnotesize{$\pm$0.02} & 87.42\footnotesize{$\pm$0.29} & 83.91\footnotesize{$\pm$0.02} & 
\underline{93.46}\footnotesize{$\pm$0.05} & 91.19\footnotesize{$\pm$0.01} & \underline{92.05}\footnotesize{$\pm$0.05} & 
96.16\footnotesize{$\pm$0.03} & \underline{95.52}\footnotesize{$\pm$0.06} & 95.05\footnotesize{$\pm$0.02} & 
\underline{83.83}\footnotesize{$\pm$0.49} & 82.15\footnotesize{$\pm$0.51} & \underline{81.74}\footnotesize{$\pm$0.72} \\
& Cross Entropy & 93.40\footnotesize{$\pm$0.20} & 92.13\footnotesize{$\pm$0.21} & 92.30\footnotesize{$\pm$0.23} & 86.99\footnotesize{$\pm$0.22} & 82.26\footnotesize{$\pm$0.75} & 84.21\footnotesize{$\pm$0.50} & 
93.05\footnotesize{$\pm$0.14} & 90.85\footnotesize{$\pm$0.37} & 91.49\footnotesize{$\pm$0.32} &
\underline{96.50}\footnotesize{$\pm$0.14} & 95.30\footnotesize{$\pm$0.13} & \underline{95.25}\footnotesize{$\pm$0.18} &
83.15\footnotesize{$\pm$0.23} & 82.63\footnotesize{$\pm$0.74} & 81.57\footnotesize{$\pm$0.30} \\
\cmidrule(lr){2-17}
& GraphSmote & 89.72\footnotesize{$\pm$0.45} & 90.69\footnotesize{$\pm$0.57} & 88.90\footnotesize{$\pm$0.48} & 85.36\footnotesize{$\pm$0.72} & 84.79\footnotesize{$\pm$1.22} & 85.22\footnotesize{$\pm$0.98} & 
87.44\footnotesize{$\pm$0.24} & 85.08\footnotesize{$\pm$0.63} & 84.26\footnotesize{$\pm$0.52} & 
95.09\footnotesize{$\pm$0.53} & 93.01\footnotesize{$\pm$0.66} & 93.42\footnotesize{$\pm$0.76} &
82.94\footnotesize{$\pm$0.70} & 80.42\footnotesize{$\pm$1.14} & 80.65\footnotesize{$\pm$0.75} \\
& GraphENS & 93.37\footnotesize{$\pm$0.42} & 92.18\footnotesize{$\pm$0.36} & 91.63\footnotesize{$\pm$0.46} & 86.35\footnotesize{$\pm$0.71} & 87.66\footnotesize{$\pm$0.54} & \underline{85.81}\footnotesize{$\pm$0.47} & 
91.65\footnotesize{$\pm$0.23} & 90.72\footnotesize{$\pm$0.39} & 89.53\footnotesize{$\pm$0.36} &
95.46\footnotesize{$\pm$0.09} & 95.32\footnotesize{$\pm$0.04} & 94.27\footnotesize{$\pm$0.06} &
81.78\footnotesize{$\pm$0.06} & 80.87\footnotesize{$\pm$0.12} & 79.80\footnotesize{$\pm$0.08} \\
& TAM & 90.13\footnotesize{$\pm$0.33} & 90.98\footnotesize{$\pm$0.36} & 89.15\footnotesize{$\pm$0.49} & 
85.46\footnotesize{$\pm$0.11} & 88.51\footnotesize{$\pm$0.67} & 84.52\footnotesize{$\pm$0.26} & 
92.41\footnotesize{$\pm$0.04} & 90.84\footnotesize{$\pm$0.01} & 91.35\footnotesize{$\pm$0.02} & 
95.35\footnotesize{$\pm$0.19} & 95.04\footnotesize{$\pm$0.08} & 94.04\footnotesize{$\pm$0.20} &
80.73\footnotesize{$\pm$0.34} & 79.02\footnotesize{$\pm$0.21} & 79.02\footnotesize{$\pm$0.16} \\
& GraphSHA & \underline{93.63}\footnotesize{$\pm$0.23} & \underline{92.61}\footnotesize{$\pm$0.66} & \underline{92.60}\footnotesize{$\pm$0.38} & 82.98\footnotesize{$\pm$0.17} & 77.73\footnotesize{$\pm$1.90} & 79.10\footnotesize{$\pm$2.22} & 
92.68\footnotesize{$\pm$0.59} & 91.00\footnotesize{$\pm$0.37} & 90.94\footnotesize{$\pm$0.51} & 
96.27\footnotesize{$\pm$0.14} & 95.51\footnotesize{$\pm$0.17} & 95.05\footnotesize{$\pm$0.11} &
81.83\footnotesize{$\pm$1.22} & 80.76\footnotesize{$\pm$0.54} & 78.95\footnotesize{$\pm$1.53} \\
& BuffGraph & \textbf{93.90}\footnotesize{$\pm$0.13} & \textbf{93.10}\footnotesize{$\pm$0.09} & \textbf{93.20}\footnotesize{$\pm$0.27} & \textbf{90.22}\footnotesize{$\pm$0.48} & \textbf{91.85}\footnotesize{$\pm$0.34} & \textbf{89.30}\footnotesize{$\pm$0.69} & 
\textbf{94.90}\footnotesize{$\pm$0.28} & \textbf{93.88}\footnotesize{$\pm$0.39} & \textbf{93.70}\footnotesize{$\pm$0.41} &
\textbf{96.78}\footnotesize{$\pm$0.07} & \textbf{96.02}\footnotesize{$\pm$0.16} & \textbf{95.79}\footnotesize{$\pm$0.07} &
\textbf{84.47}\footnotesize{$\pm$0.22} & \textbf{83.62}\footnotesize{$\pm$0.15} & \textbf{82.10}\footnotesize{$\pm$0.12} \\
\bottomrule
\end{tabular}
}
\end{center}
\vspace{-0.1cm}
\end{table*}

\begin{table*}[!t]
\centering
\begin{center}
\caption{Imbalanced splitting experiment results of our model BuffGraph and other baselines on five class-imbalanced node classification benchmark datasets. }\label{table:imbalanced-results}
\vspace{-0.1cm}
\scalebox{0.65}{
\begin{tabular}{cl|ccc|ccc|ccc|ccc|ccc}
\toprule
\multirow{2}{*}{} & \textbf{Dataset} & \multicolumn{3}{c|}{Amazon-Photos} & \multicolumn{3}{c|}{Amazon-Computers} & \multicolumn{3}{c|}{Coauthor-CS} & \multicolumn{3}{c|}{Coauthor-Physics} & \multicolumn{3}{c}{WikiCS} \\
\cmidrule(lr){3-5}\cmidrule(lr){6-8}\cmidrule(lr){9-11}\cmidrule(lr){11-13}\cmidrule(lr){13-15}\cmidrule(lr){15-17}
& $\rho$=10 & {Acc.} & {BAcc.} & {F1} & {Acc.} & {BAcc.} & {F1} & {Acc.} & {BAcc.} & {F1} & {Acc.} & {BAcc.} & {F1} & {Acc.} & {BAcc.} & {F1}  \\
\midrule
\multirow{11}{*}{\rotatebox{90}{\textbf{Methods}}}
& Vanilla & 92.20\footnotesize{$\pm$0.54} & 89.60\footnotesize{$\pm$0.05} & 90.41\footnotesize{$\pm$0.14} & 
83.40\footnotesize{$\pm$0.29} & 69.71\footnotesize{$\pm$0.28} & 70.79\footnotesize{$\pm$0.43} & 
92.54\footnotesize{$\pm$0.55} & 89.86\footnotesize{$\pm$0.68} & 90.53\footnotesize{$\pm$0.63} &
95.65\footnotesize{$\pm$0.04} & 93.76\footnotesize{$\pm$0.12} & 94.19\footnotesize{$\pm$0.17} & 
81.30\footnotesize{$\pm$1.00} & 75.16\footnotesize{$\pm$1.53} & 77.42\footnotesize{$\pm$1.55} \\
\cmidrule(lr){2-17}
& Reweight & 92.65\footnotesize{$\pm$0.36} & 92.34\footnotesize{$\pm$0.17} & 90.79\footnotesize{$\pm$0.36} & 86.46\footnotesize{$\pm$0.20} & 89.26\footnotesize{$\pm$0.08} & 85.33\footnotesize{$\pm$0.14} & 
93.23\footnotesize{$\pm$0.12} & 91.74\footnotesize{$\pm$0.07} & 91.86\footnotesize{$\pm$0.03} &
96.35\footnotesize{$\pm$0.04} & 95.12\footnotesize{$\pm$0.20} & 95.16\footnotesize{$\pm$0.09} &
81.16\footnotesize{$\pm$0.13} & 81.48\footnotesize{$\pm$0.28} & 79.54\footnotesize{$\pm$0.34}\\
& PC Softmax & 84.51\footnotesize{$\pm$0.86} & 88.69\footnotesize{$\pm$1.27} & 84.01\footnotesize{$\pm$2.57} & 70.48\footnotesize{$\pm$1.09} & 84.92\footnotesize{$\pm$1.21} & 70.50\footnotesize{$\pm$0.46} & 
92.78\footnotesize{$\pm$0.02} & 93.16\footnotesize{$\pm$0.06} & 91.23\footnotesize{$\pm$0.14} &
95.18\footnotesize{$\pm$0.09} & \textbf{95.47}\footnotesize{$\pm$0.08} & 93.87\footnotesize{$\pm$0.13} &
76.01\footnotesize{$\pm$2.24} & 80.30\footnotesize{$\pm$1.42} & 73.85\footnotesize{$\pm$2.13} \\
& Balanced Softmax & 92.81\footnotesize{$\pm$0.20} & \textbf{93.33}\footnotesize{$\pm$0.04} & 91.44\footnotesize{$\pm$0.19} & \underline{87.55}\footnotesize{$\pm$0.24} & \underline{89.31}\footnotesize{$\pm$0.16} & \underline{86.95}\footnotesize{$\pm$0.02} & 
93.99\footnotesize{$\pm$0.01} & \underline{93.24}\footnotesize{$\pm$0.03} & 92.35\footnotesize{$\pm$0.02} & 
\textbf{96.46}\footnotesize{$\pm$0.05} & \underline{95.46}\footnotesize{$\pm$0.03} & \underline{95.32}\footnotesize{$\pm$0.04} & 
82.14\footnotesize{$\pm$0.04} & \underline{82.45}\footnotesize{$\pm$0.21} & \underline{80.10}\footnotesize{$\pm$0.08} \\
& Cross Entropy & 91.67\footnotesize{$\pm$0.16} & 87.85\footnotesize{$\pm$0.40} & 89.93\footnotesize{$\pm$0.35} & 87.46\footnotesize{$\pm$0.18} & 83.49\footnotesize{$\pm$0.53} & 85.19\footnotesize{$\pm$0.53} & 
\underline{94.04}\footnotesize{$\pm$0.07} & 92.03\footnotesize{$\pm$0.03} & \underline{92.38}\footnotesize{$\pm$0.04} &
96.12\footnotesize{$\pm$0.01} & 94.53\footnotesize{$\pm$0.11} & 94.93\footnotesize{$\pm$0.03} &
82.44\footnotesize{$\pm$0.23} & 78.07\footnotesize{$\pm$0.51} & 80.06\footnotesize{$\pm$0.10} \\
\cmidrule(lr){2-17}
& GraphSmote & 88.31\footnotesize{$\pm$0.63} & 88.15\footnotesize{$\pm$1.53} & 87.27\footnotesize{$\pm$0.28} & 85.30\footnotesize{$\pm$0.66} & 84.66\footnotesize{$\pm$0.27} & 84.35\footnotesize{$\pm$0.23} & 
88.95\footnotesize{$\pm$0.19} & 83.96\footnotesize{$\pm$0.99} & 85.56\footnotesize{$\pm$0.78} & 
92.64\footnotesize{$\pm$0.36} & 92.79\footnotesize{$\pm$0.11} & 94.42\footnotesize{$\pm$0.53} &
74.96\footnotesize{$\pm$1.07} & 69.43\footnotesize{$\pm$2.17} & 70.82\footnotesize{$\pm$1.93} \\
& GraphENS & 92.55\footnotesize{$\pm$0.07} & 91.66\footnotesize{$\pm$0.37} & 91.07\footnotesize{$\pm$0.02} & 85.50\footnotesize{$\pm$0.58} & 89.21\footnotesize{$\pm$0.14} & 85.05\footnotesize{$\pm$0.69} & 
92.12\footnotesize{$\pm$0.03} & 90.49\footnotesize{$\pm$0.01} & 89.21\footnotesize{$\pm$0.16} &
95.35\footnotesize{$\pm$0.19} & 95.04\footnotesize{$\pm$0.08} & 94.04\footnotesize{$\pm$0.20} &
80.73\footnotesize{$\pm$0.30} & 79.94\footnotesize{$\pm$0.27} & 77.83\footnotesize{$\pm$0.45} \\
& TAM & 91.08\footnotesize{$\pm$0.03} & 91.70\footnotesize{$\pm$0.07} & 90.15\footnotesize{$\pm$0.07} & 
85.79\footnotesize{$\pm$0.18} & 88.21\footnotesize{$\pm$0.69} & 85.21\footnotesize{$\pm$0.42} & 
92.53\footnotesize{$\pm$0.04} & 90.45\footnotesize{$\pm$0.13} & 90.67\footnotesize{$\pm$0.13} & 
95.25\footnotesize{$\pm$0.06} & 94.11\footnotesize{$\pm$0.19} & 93.66\footnotesize{$\pm$0.16} &
81.12\footnotesize{$\pm$0.04} & 80.29\footnotesize{$\pm$0.03} & 79.32\footnotesize{$\pm$0.12} \\
& GraphSHA & \underline{93.56}\footnotesize{$\pm$0.04} & 92.46\footnotesize{$\pm$0.30} & \underline{92.59}\footnotesize{$\pm$0.02} & 85.24\footnotesize{$\pm$0.52} & 83.77\footnotesize{$\pm$0.55} & 83.31\footnotesize{$\pm$0.59} & 
92.42\footnotesize{$\pm$0.16} & 90.43\footnotesize{$\pm$0.46} & 90.21\footnotesize{$\pm$0.22} & 
96.27\footnotesize{$\pm$0.14} & 95.14\footnotesize{$\pm$0.04} & 94.94\footnotesize{$\pm$0.14} &
\underline{82.60}\footnotesize{$\pm$0.30} & 80.34\footnotesize{$\pm$0.31} & 80.00\footnotesize{$\pm$0.52} \\
& BuffGraph & \textbf{93.74}\footnotesize{$\pm$0.30} & \underline{92.90}\footnotesize{$\pm$0.42} & \textbf{92.65}\footnotesize{$\pm$0.60} & \textbf{89.51}\footnotesize{$\pm$0.35} & \textbf{90.54}\footnotesize{$\pm$0.57} & \textbf{88.14}\footnotesize{$\pm$0.40} & 
\textbf{94.06}\footnotesize{$\pm$0.02} & \textbf{93.78}\footnotesize{$\pm$0.03} & \textbf{92.72}\footnotesize{$\pm$0.35} &
\underline{96.43}\footnotesize{$\pm$0.17} & 95.32\footnotesize{$\pm$0.04} & \textbf{95.33}\footnotesize{$\pm$0.12} &
\textbf{83.71}\footnotesize{$\pm$0.55} & \textbf{83.00}\footnotesize{$\pm$0.52} & \textbf{81.93}\footnotesize{$\pm$0.62} \\
\bottomrule
\end{tabular}
}
\end{center}
\vspace{-0.1cm}
\end{table*}

\subsection{Experiments Setup}
\textbf{Environments.} We conduct all experiments using PyTorch on an NVIDIA GeForce RTX 3090 GPU. More details are in \ref{appendix:env}.

\noindent \textbf{Datasets.} To comprehensively evaluate BuffGraph, we conduct experiments across five naturally class-imbalanced datasets: Amazon-Photos, Amazon-Computers, Coauthor-CS, Coauthor-Physics, and WikiCS. A statistical summary of these datasets, including their key characteristics, is detailed in Table \ref{table:datasets}. The Max/Min ratio in this summary represents the number of samples in the largest majority class to that in the smallest minority class, offering insights into the extent of imbalance within each dataset.

\noindent \textbf{Baselines.} In this study, we present our results using a GCN backbone and compare them against a comprehensive set of baselines that encompass both loss management strategies and class-imbalanced node classification approaches. These include Reweight, PC SoftMax, Cross Entropy, and Balanced Softmax for loss management, alongside GraphSMOTE, GraphENS, TAM, and GraphSHA for tackling class imbalance. A detailed description of each baseline is provided in the Appendix \ref{appendix:baseline}.
    

\noindent \textbf{Evaluation Metrics. } We evaluate model performance using Accuracy (Acc.), Balanced Accuracy (BAcc.), and Macro F1 Score (F1). Balanced Accuracy is defined as the average of accuracy of all classes \cite{li2023graphsha}. Macro F1 is defined as the average of F1 scores of all classes. These metrics are selected for their widespread acceptance in class-imbalanced node classification task \cite{zhao2021graphsmote, park2022graphens, song2022tam, zhou2023graphsr}.

\subsection{Implementation Details}
\textbf{Random Splitting Experiment Settings.}
For the random splitting experiment, we divide the datasets into training, validation, and testing sets with a 6:2:2 ratio. This random partitioning is carefully chosen to preserve the inherent distributional characteristics of each dataset, thereby mitigating potential biases. It ensures that our evaluations accurately reflect the true learning capabilities of the models and not artifacts of the data distribution. We standardize our models on the three hidden layers and explore hidden dimensions of 64, 128, and 256. Among these, we select the dimension that provides the best performance. For the learning rate, we explore the optimal settings by evaluating the performance at values of 0.001, 0.005, and 0.01. For the dropout rate, we explore the optimal settings for evaluating the performance of 0.1, 0.2, 0.3, 0.4, 0.5. Based on these experiments, we have chosen to set the hidden dimensions to 256, the dropout rate to 0.4, and the learning rate to 0.01 to achieve the best performance. Regarding the training duration, we configure the model to run for up to 2000 epochs, with an early stopping parameter set at 500 epochs to ensure stability in the model's predictions during the training phase.

\noindent \textbf{Imbalanced Experiment Settings.}
For the imbalanced experiment, we initially perform a random split of the dataset into training, validation, and testing sets following a 6:2:2 ratio. Subsequently, within the training set, we impose an imbalance by downsampling the last half of the classes to achieve an imbalance ratio of 10, meaning that the number of samples in these minority classes is reduced to one-tenth of the most samples in the majority classes. This two-step process, which adheres to the guidelines in \cite{park2022graphens}, allows us to simulate a realistic imbalanced learning scenario and rigorously assess model performance under the imbalanced condition.

\noindent \textbf{Baseline Settings:} In our experiments, we meticulously adhere to the hyperparameter settings outlined in the papers for baseline models GraphSMOTE \cite{zhao2021graphsmote}, GraphENS \cite{park2022graphens}, TAM \cite{song2022tam}, and GraphSHA \cite{li2023graphsha}. This approach is critical to ensure the reproducibility and comparability of our study. More details are in Appendix~\ref{appendix:baseline}.




\subsection{Experimental Results}

\textbf{Random Splitting Experiment Results.} 
The results presented in Table \ref{table:main-results} highlight BuffGraph's outstanding performance against competing methods across all metrics and datasets with a notable advantage in BAcc. For example, BuffGraph exhibits a 2\% increase in BAcc over the next best performing model on Amazon-Computers and Coauthor-CS. This enhancement underscores BuffGraph's proficiency in accurately classifying minority classes. Additionally, BuffGraph excels in overall accuracy, affirming its adeptness at boosting performance uniformly across both majority and minority classes without compromising the integrity of either.

The exceptional performance of BuffGraph can be attributed to its innovative incorporation of buffer nodes, which strategically modulate the impact of majority class nodes. This prevents the excessive influence of majority class on minority class nodes. This approach ensures a more balanced information dissemination within the graph, allowing minority class nodes to retain their distinctive features during the learning process. In contrast, while some baselines, such as Reweight, achieve impressive BAcc. on specific datasets, their inability to maintain high overall accuracy suggests a disproportionate focus on minority classes at the expense of majority class representation. This imbalance highlights the challenges inherent in designing models that can navigate the complex dynamics of class-imbalanced datasets without favoring one class group over another. Therefore, BuffGraph represents a significant step forward in achieving a balance between accurate minority class prediction and overall classification performance.


\noindent \textbf{Imbalanced Splitting Experiment Results.} 
Different to the random splitting setting in the above experiment, we specifically focused on cases where the imbalance ratio between majority classes and minority classes equals 10 to test BuffGraph's performance against other baselines in more extreme class-imbalanced graphs. From the result shown in Table \ref{table:imbalanced-results}, BuffGraph has superior performance across most metrics and most datasets, particularly in terms of F1. For example, BuffGraph achieves a 2\% increase in F1 on Amazon-Computers and WikiCS than runner-up results. These findings highlight BuffGraph's adeptness at sustaining elevated performance levels amidst substantial class imbalances. Consequently, BuffGraph is affirmed as a robust model, adept not just in randomly class-imbalanced graphs but also in those with significant class imbalances as defined in our study.

\vspace{-0.2cm}
\begin{figure}[!t]
\centering
\includegraphics[width=0.9\linewidth]{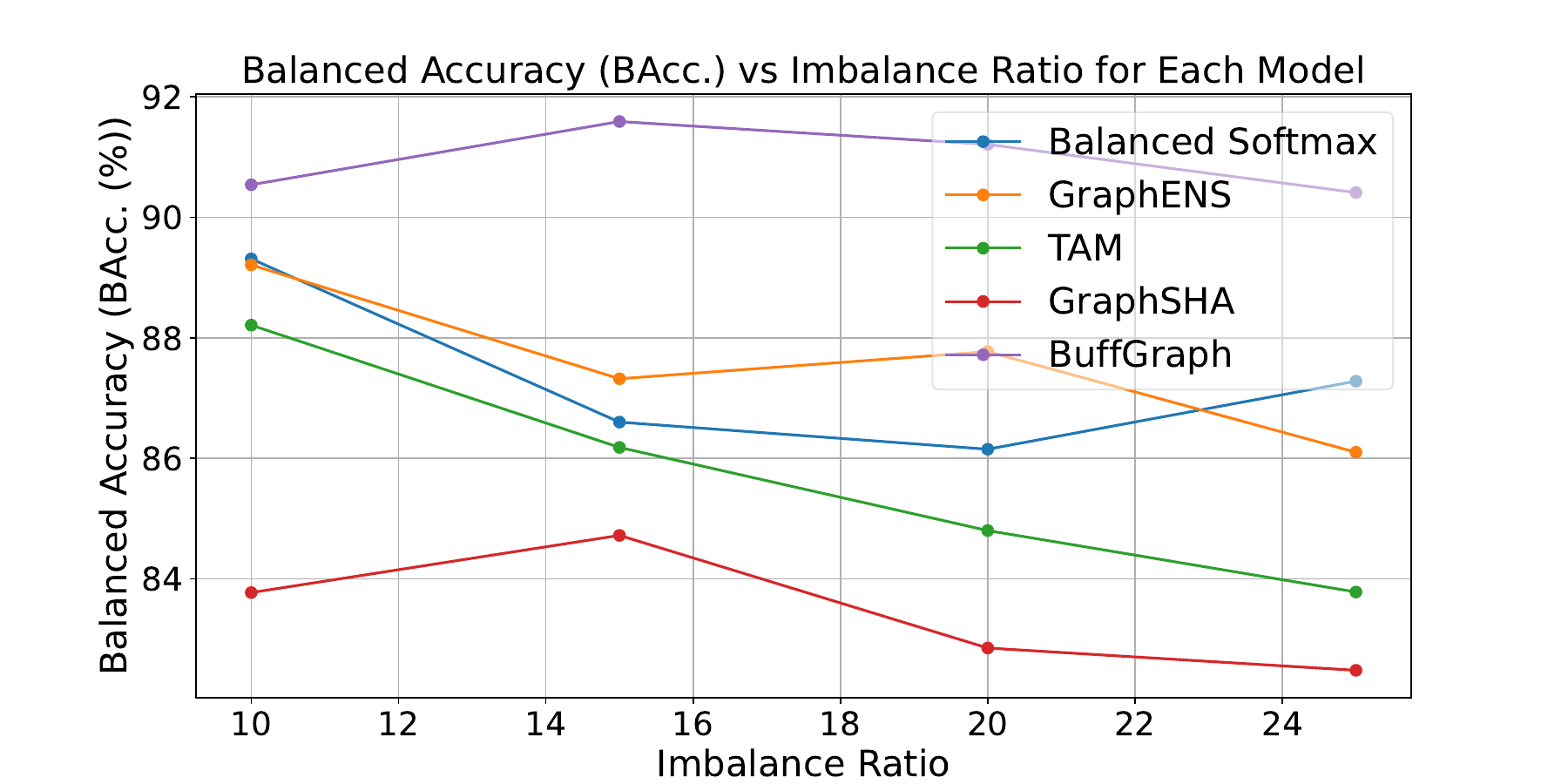}
\centering
\caption{Trend of BAcc. with the increase of imbalance ratio on Amazon-Computers.}
\label{figure:imbalance_ratio}
\vspace{-0.4cm}
\end{figure}

\subsection{Imbalance Ratio Study}
To further investigate BuffGraph's effectiveness on class-imbalanced graphs, we adjust the imbalance ratio to 15, 20, and 25 within the Amazon-Computers dataset. Subsequently, we evaluate the BAcc. against competitive baselines, including Balanced Softmax, GraphENS, TAM, GraphSHA, as depicted in Figure \ref{figure:imbalance_ratio}. BuffGraph demonstrates stable performance across varying imbalance ratios, significantly surpassing other baselines by at least a 3\% margin in comparison to the second-best results. This underscores BuffGraph's robustness in managing class imbalance across different levels of imbalance ratios for node classification tasks in class-imbalanced graphs.


\vspace{-0.2cm}
\begin{figure}[!t]
\centering
\includegraphics[width=0.9\linewidth]{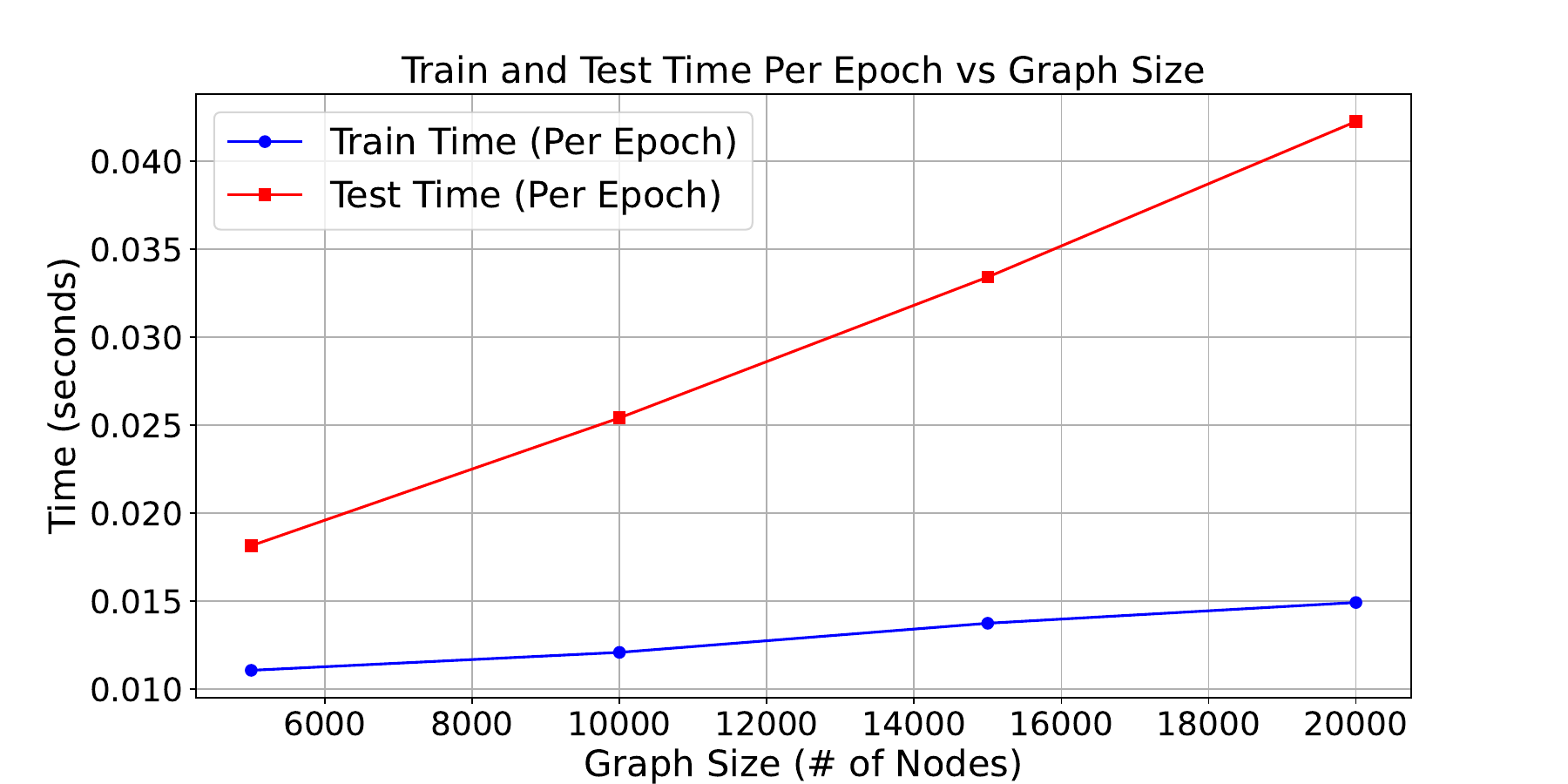}
\centering
\caption{Scalability study on Coauthor-Physics.}
\label{figure:scalability}
\vspace{-0.4cm}
\end{figure}

\subsection{Scalability Study}
To evaluate the scalability of BuffGraph, we conduct a scalability study on the biggest dataset: Coauthor-Physics. We sample four graphs of Coauthor-Physics with the number of nodes: 5000, 10000, 15000, and 20000. Then, we run BuffGraph on these graphs and report the training and test time of each epoch. We report the result in the Figure \ref{figure:scalability}. We can see from the Figure \ref{figure:scalability} that the time of each epoch increases linearly with the sample of the graph's node size which shows that BuffGraph can scale to large graphs in the real-world scenarios. 

\begin{figure}[!t]
\centering
\includegraphics[width=0.9\linewidth]{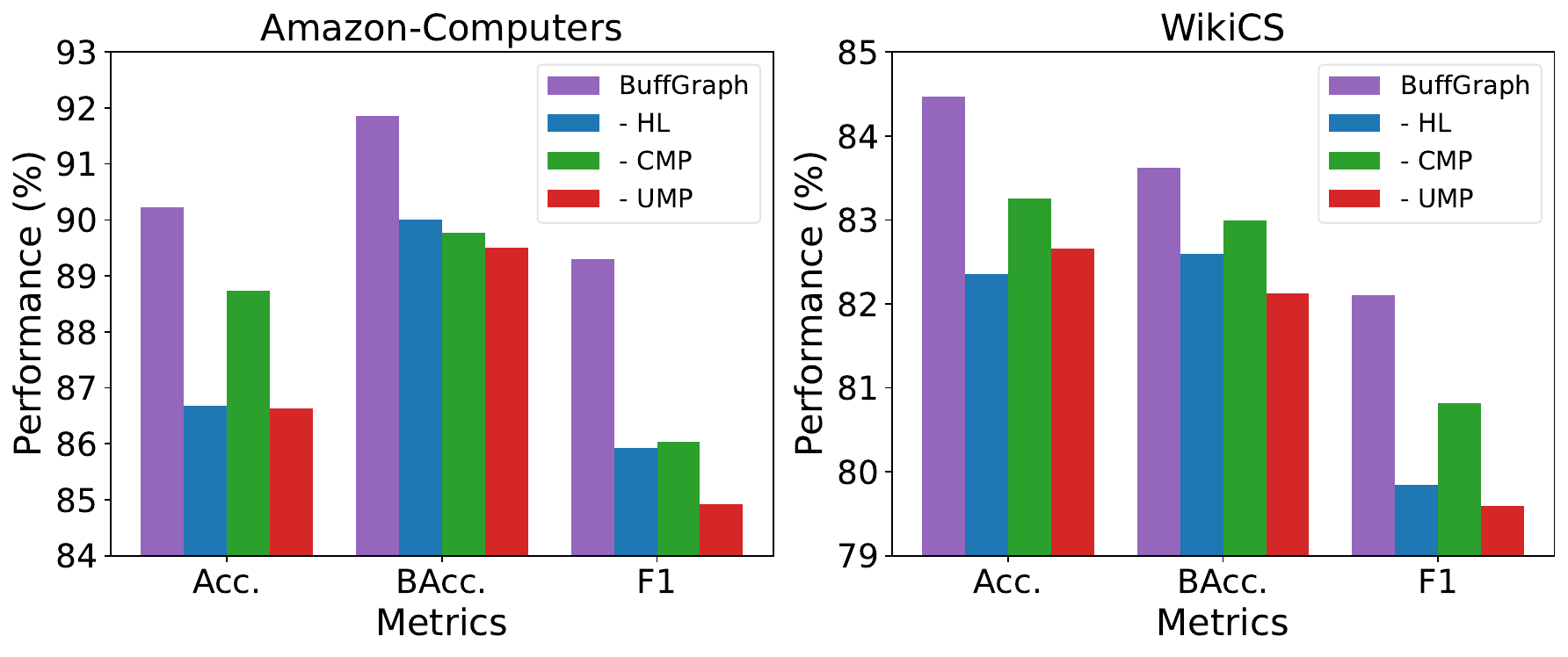}
\centering
\caption{Ablation study on Amazon-Computers and WikiCS. }
\label{figure:ablation}
\vspace{-0.4cm}
\end{figure}


\subsection{Ablation Study} 
To assess the efficacy of each element within our proposed BuffGraph model, we conduct an ablation study on the Amazon-Computers and WikiCS datasets, applying the GCN architecture and a random splitting setting, which is of the same setting as the experiment in Table \ref{table:main-results}. This study deconstructs the BuffGraph model by removing its key components once at a time to understand their individual contributions. The variations considered in this analysis include: (1) BuffGraph, representing the full BuffGraph model as the baseline for comparison; (2) BuffGraph - Heterophily Loss (HL), examining the effect of removing the heterophily loss component on overall performance; (3) BuffGraph - Concrete Message Passing (CMP), assessing the role of concrete message passing and its dependency on the buffer node path; and (4) BuffGraph - Update Message Passing (UMP), investigating the implications of excluding the update of message passing during the training based on total loss. This structured approach allows for a nuanced understanding of how each component contributes to the efficacy of the BuffGraph model.

The results shown in Figure \ref{figure:ablation} underscore the critical nature of the updating message passing mechanism in training, as its removal (- UMP) precipitates the most substantial decline in performance across most metrics on the both datasets, especially of F1. In addition, the exclusion of other components also notably diminishes all metrics across both datasets. This analysis underscores the integral role each component plays within the BuffGraph model, highlighting their collective importance in achieving optimal performance.


\begin{figure}[!t]
\centering
\includegraphics[width=0.9\linewidth]{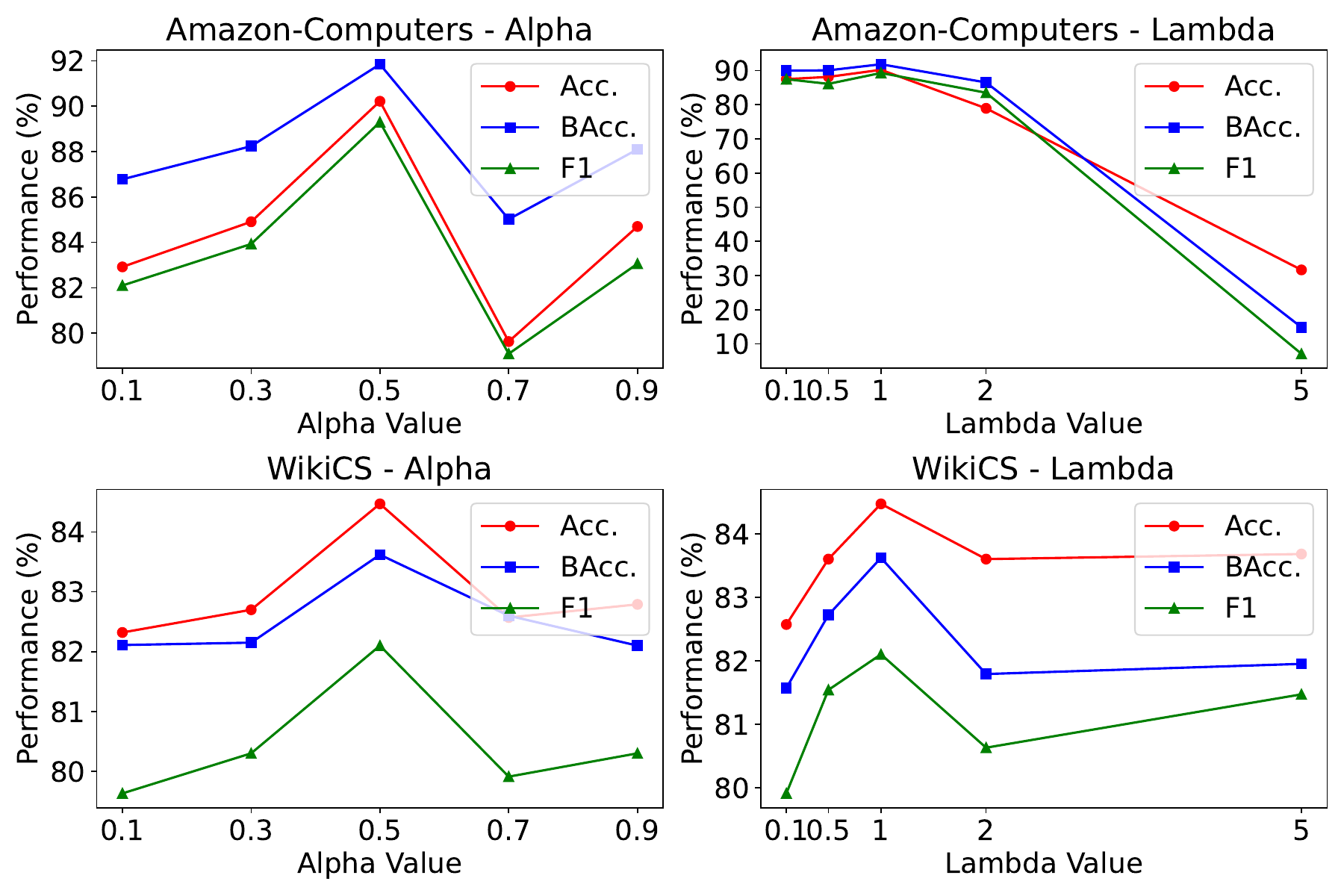}
\centering
\caption{Parameter sensitivity analysis on Amazon-Computers and WikiCS.}
\label{figure:parameters}
\vspace{-0.4cm}
\end{figure}

\subsection{Parameter Sensitivity Study} \label{sec:parameter}
To assess the sensitivity of BuffGraph's parameters, we focus on two critical hyperparameters: $\alpha$, utilized in feature mixup, and $\lambda$, applied in heterophily loss management. This investigation employs Amazon-Computers and WikiCS. The outcomes are detailed in Figure \ref{figure:parameters}. Notably, a $\alpha$ value of 0.5 yields the most favorable results across both datasets. Regarding $\lambda$, performance on Amazon-Computers remains relatively stable for values up from 0.1 to 1, whereas on WikiCS, performance improves as $\lambda$ increases from 0.1 to 1. However, elevating $\lambda$ to 2 and 5 lead to a marked decline in performance for both datasets. This parameter sensitivity study highlights the nuanced impact of these parameters on model efficacy as well as the wise choice of these parameters in our BuffGraph.

%% file: sec-conclusion.tex
\section{Conclusion}
In this study, we introduce BuffGraph, a groundbreaking model designed to address the challenges of class imbalance in graph-structured data, with a particular emphasis on heterophily. Leveraging a novel buffer node interpolation technique and an adaptive message-passing mechanism, BuffGraph significantly surpasses existing baselines in the realm of class-imbalanced node classification. Our comprehensive experimental evaluation underscores BuffGraph's exceptional proficiency in accurately classifying minority classes while preserving the integrity of majority class identification across both naturally and artificially imbalanced graphs. This capability is vital for a broad spectrum of practical applications. The insights gained from our study not only affirm the practical utility of BuffGraph but also open new avenues for further research into the complex dynamics of heterophily in class-imbalanced graph analysis.

%% file: sec-appendix.tex
\clearpage
\section*{APPENDIX}

\section{EXPERIMENTAL ENVIRONMENTS}\label{appendix:env}
All models in our experiments were implemented using Pytorch 2.0.0 in Python 3.9.16, and run on a robust Linux workstation. This system is equipped with two Intel(R) Xeon(R) Gold 6226R CPUs, each operating at a base frequency of 2.90 GHz and a max turbo frequency of 3.90 GHz. With 16 cores each, capable of supporting 32 threads, these CPUs offer a total of 64 logical CPUs for efficient multitasking and parallel computing. The workstation is further complemented by a potent GPU setup, comprising eight NVIDIA GeForce RTX 3090 GPUs, each providing 24.576 GB of memory. The operation of these GPUs is managed by the NVIDIA-SMI 525.60.13 driver and CUDA 12.0, ensuring optimal computational performance for our tasks.

\section{DETAILED EXPERIMENT SETTINGS}\label{appendix:baseline}
For the class-imbalanced node classification experiments, we selected a diverse set of competitive baselines, categorized into loss management strategies and specific approaches to addressing class imbalance in node classification:

\textbf{Loss Management Strategies:}
\begin{itemize}
    \item \textit{Reweight}: Modifies class weights inversely proportional to their frequency in the dataset, aiming to mitigate class imbalance.
    \item \textit{PC SoftMax} \cite{hong2021disentangling}: Enhances model probability calibration for multi-class scenarios, improving minority class predictions.
    \item \textit{Cross Entropy}: Employed as the primary loss function, it quantifies the difference between the predicted probabilities and the actual distribution.
    \item \textit{Balanced Softmax} \cite{ren2020balanced}: Adjusts softmax regression to lower the generalization error, particularly effective in multi-class imbalance.
\end{itemize}

\textbf{Class-Imbalanced Node Classification Approaches:}
\begin{itemize}
    \item \textit{GraphSMOTE} \cite{zhao2021graphsmote}: Generates synthetic instances for minority classes, directly addressing the imbalance issue.
    \item \textit{GraphENS} \cite{park2022graphens}: Employs ensemble techniques to augment minority class representation through synthetic instance generation.
    \item \textit{TAM} \cite{song2022tam}: Introduces a tailored margin that considers connectivity and distribution, enhancing minority class classification.
    \item \textit{GraphSHA} \cite{li2023graphsha}: Focuses on creating challenging minority samples to improve classification margins between classes.
\end{itemize}

We employ the same GCN backbone \cite{kipf2016semi}, for all class-imbalanced node classification baselines. Typically, the selection of the number of hidden layers is confined to the set \{2, 3\}, and the dimensions of these layers are chosen from \{64, 128, 256\}. We report the best prediction results obtained from all configurations.

For GraphSmote, we opt for the GraphSmote\(_O\) variant, which is tailored to predict discrete values without necessitating pretraining, showcasing superior performance among its various versions \cite{zhao2021graphsmote}.

In the implementation of GraphENS, we adhere to the configurations suggested in the original codebase, setting the feature masking rate \(k\) to 0.01 and the temperature \(\tau\) to 1 \cite{park2022graphens}.

For TAM, we select the GraphENS-based iteration, which is identified as the most performant according to the findings reported in the corresponding paper. The default settings from the released code are utilized, with the coefficients for ACM \(\alpha\), ADM \(\beta\), and classwise temperature \(\phi\) set to 2.5, 0.5, and 1.2, respectively \cite{song2022tam}.

Specifically for GraphSHA, we follow the parameter configurations detailed in the original study, employing the PPR version with a setting of \(\alpha=0.05\) and \(K=128\) \cite{li2023graphsha}.